\documentclass[10pt,twocolumn,letterpaper]{article}

\usepackage[pagenumbers]{cvpr} % To force page numbers, e.g. for an arXiv version

\usepackage{graphicx}
\usepackage{amsmath}
\usepackage{amssymb}
\usepackage{booktabs}

\usepackage[pagebackref,breaklinks,colorlinks]{hyperref}

\usepackage[capitalize]{cleveref}
\crefname{section}{Sec.}{Secs.}
\Crefname{section}{Section}{Sections}
\Crefname{table}{Table}{Tables}
\crefname{table}{Tab.}{Tabs.}

\def\cvprPaperID{*****} % *** Enter the CVPR Paper ID here
\def\confName{CVPR}
\def\confYear{2023}

\begin{document}

\title{A Critical Look at the Current Usage of Foundation Model for Dense Recognition Task
}

\author{Shiqi Yang$^{1,2}$\thanks{Work is done during intern at OMRON SINIC X.}, Atsushi Hashimoto$^{3}$, Yoshitaka Ushiku$^{3}$\\
$^{1}$ Computer Vision Center, Bellaterra, Spain\\
$^{2}$ Department of Computer Science, Universitat Autònoma de Barcelona, Bellaterra, Spain\\
$^{3}$ OMRON SINIC X, Tokyo, Japan\\
{\tt\small syang@cvc.uab.es}, \tt\small{\{atsushi.hashimoto, yoshitaka.ushiku\}@sinicx.com}
}

\maketitle

%%%%%%%%% ABSTRACT
\begin{abstract}
   In recent years large model trained on huge amount of cross-modality data, which is usually be termed as foundation model, achieves conspicuous accomplishment in many fields, such as image recognition and generation. Though achieving great success in their original application case, it is still unclear whether those foundation models can be applied to other different downstream tasks. In this paper, we conduct a short survey on the current methods for discriminative dense recognition tasks, which are built on the pretrained foundation model. And we also provide some preliminary experimental analysis of an existing open-vocabulary segmentation method based on Stable Diffusion, which indicates the current way of deploying diffusion model for segmentation is not optimal. This aims to provide insights for future research on adopting foundation model for downstream task.
\end{abstract}

%%%%%%%%% BODY TEXT
\section{Introduction}
\label{sec:intro}

In the last decades, deep model trained with large amount of labeled data succeeds to be top-rank in almost all computer vision tasks. Besides the achievements in the supervised learning tasks, other research lines improve the generalization and universality ability, such as self-supervised learning~\cite{he2020momentum, grill2020bootstrap,chen2020simple} which empowers the model with strong representation feature learning capacity with only unlabeled data, open-set or open-world learning which endows the model with the ability to either reject~\cite{oza2019c2ae,yoshihashi2019classification,vazeopen} or distinguish~\cite{cao2021open,rizve2022openldn,vaze2022generalized,jia2021joint} novel categories, and domain generalization~\cite{gulrajani2020search,li2018domain,robey2021model} or domain adaptation~\cite{saito2018maximum,liang2020we,yang2021generalized} which improves model's generalization to test data of different distributions, to name a few.

More recently, the training of models with abundant cross modality data is becoming more popular. For example, CLIP~\cite{radford2021learning} is a visual-language model trained with huge amount of image and text pairing data, via a contrastive learning objective. Due to the learned image-language pairing representation, with the provided text prompts during inference time, the model excels at zero-shot recognition. SAM~\cite{kirillov2023segment} is a general category-agnostic segmentation/localization solution which supports several types of prompts, it is capable of segmenting whole objects or object parts of any shape. ImageBind~\cite{girdhar2023imagebind} learns a joint embedding space across six different modalities, with visual space as the intermedia embedding space, and it is a strong pipeline for cross-modality recognition tasks.

\begin{figure*}[tbp]
  \centering
  \includegraphics[width=0.9\linewidth]{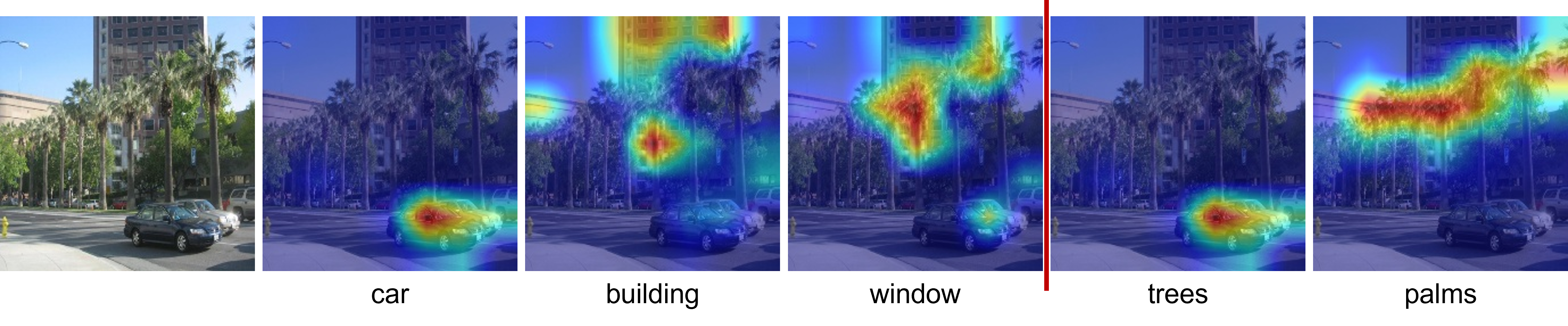}\vspace{-4mm}
  \caption{Grad-CAM visualization of pretrained CLIP visual encoder. Only four classes (used for text prompt) are considered: car, building, windows and trees/palms. \vspace{-4mm}}
  \label{fig:clip_cam}
\end{figure*}

\begin{figure}[tbp]
  \centering
  \includegraphics[width=0.9\linewidth]{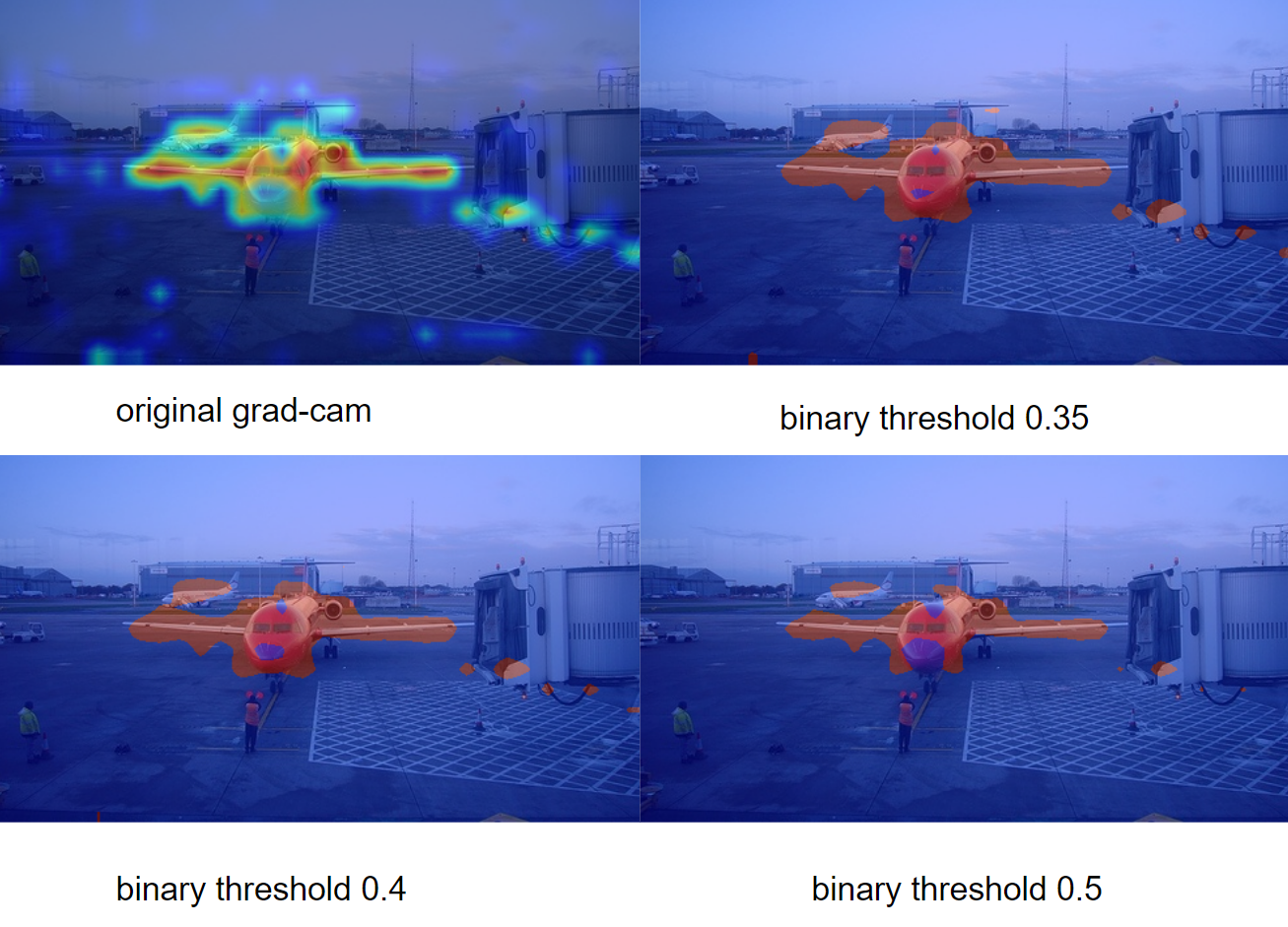}\vspace{-4mm}
  \caption{Adopting binary threshold on the Grad-CAM.\vspace{-4mm}}
  \label{fig:clip_cam_threshold}
\end{figure}

Besides large model for discriminative tasks, diffusion based\footnote{In this report, we regard (text-to-image) diffusion model also as a kind of foundation model.} image generation is another emerging hot research topic. Stable Diffusion~\cite{rombach2022high} is one of the most popular methods in both academic and non-academic communities. The pretrained Stable Diffusion could be easily adapted to the personalized data, for both image generation or editing, by fine-tuning part of the model~\cite{ruiz2023dreambooth,hu2021lora} or conducting some processing in the fixed model~\cite{hertz2022prompt,gal2022image}. Originally designed for text-to-image generation task, it can be easily extended to other conditional image generation task~\cite{zhang2023adding}, such as depth-to-image and sketch-to-image generation/translation.

With the popularity of those foundation models, a natural question arises: can those pretrained models, which are originally for image recognition or generation, be applied to other downstream tasks? As these models are trained with huge amounts of data and possess strong zero-shot recognition ability or good feature representation, the learned knowledge is expected to also facilitate other downstream tasks. This provides the possibility of using a unified model for different tasks, which could have high practical value in real-world applications. In this paper, we conduct a short survey on utilizing pretrained foundation model for downstream tasks. We mainly focus on the segmentation task, since segmentation information is also useful for other tasks such as detection and localization.

%-------------------------------------------------------------------------
%\section{Foundation Model}

\begin{figure*}[tbp]
  \centering
  \includegraphics[width=0.9\linewidth]{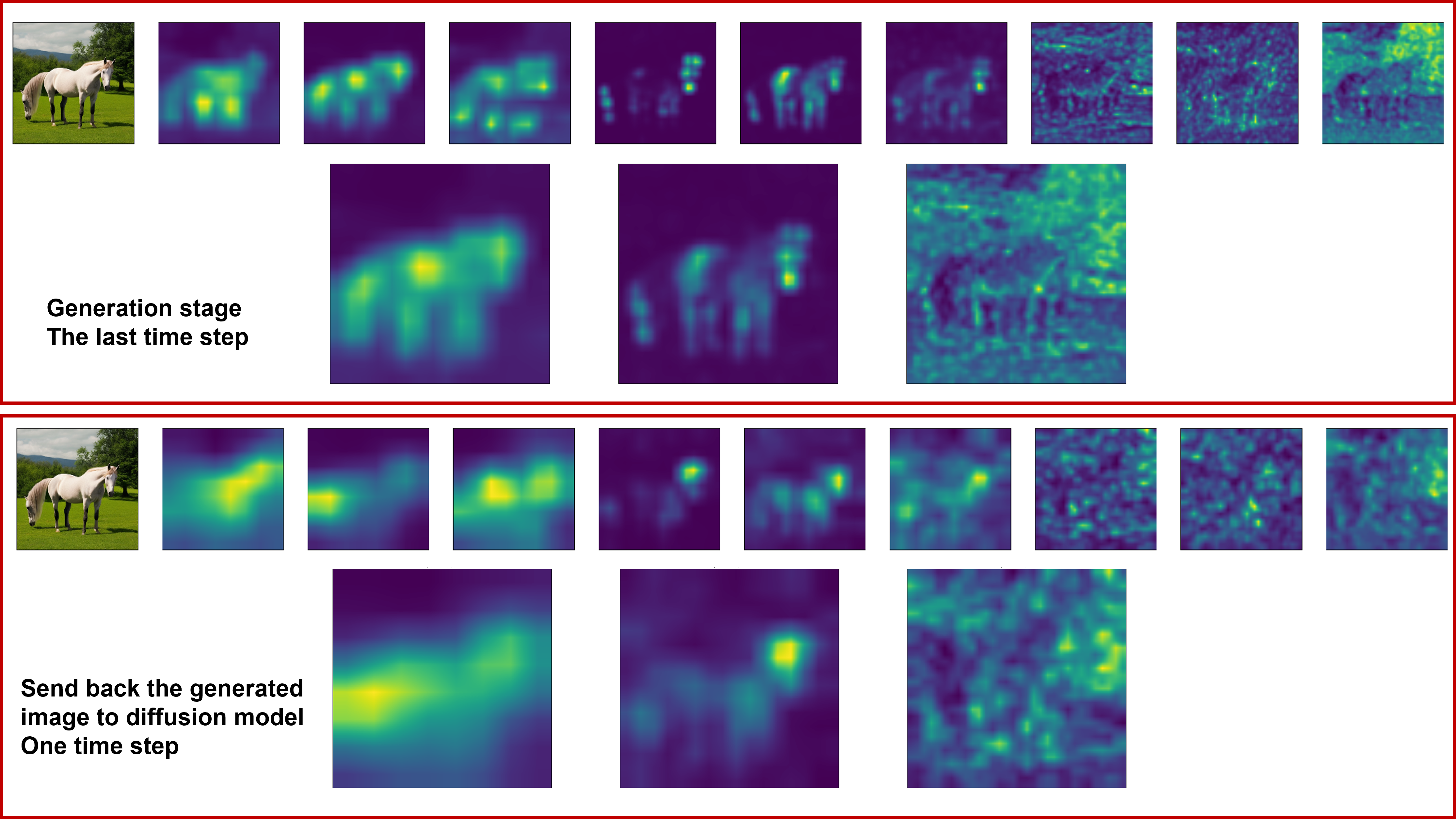}\vspace{-2mm}
  \caption{Visualization of cross attention (for token 'horse' with the text prompt 'a horse on the grass'). The upper part is the cross attention in the last timestep during generation stage. Then just like ODISE, we sent the generated image back to diffusion model with adding one noise, the visualization is shown in the lower part.  In each case, the first row shows the attention map from different layer in the output block and the second row shows the average of those attention maps of different sizes.\vspace{-2mm}}
  \label{fig:cross_atten}
\end{figure*}

\begin{figure}[tbp]
  \centering
  \includegraphics[width=0.99\linewidth]{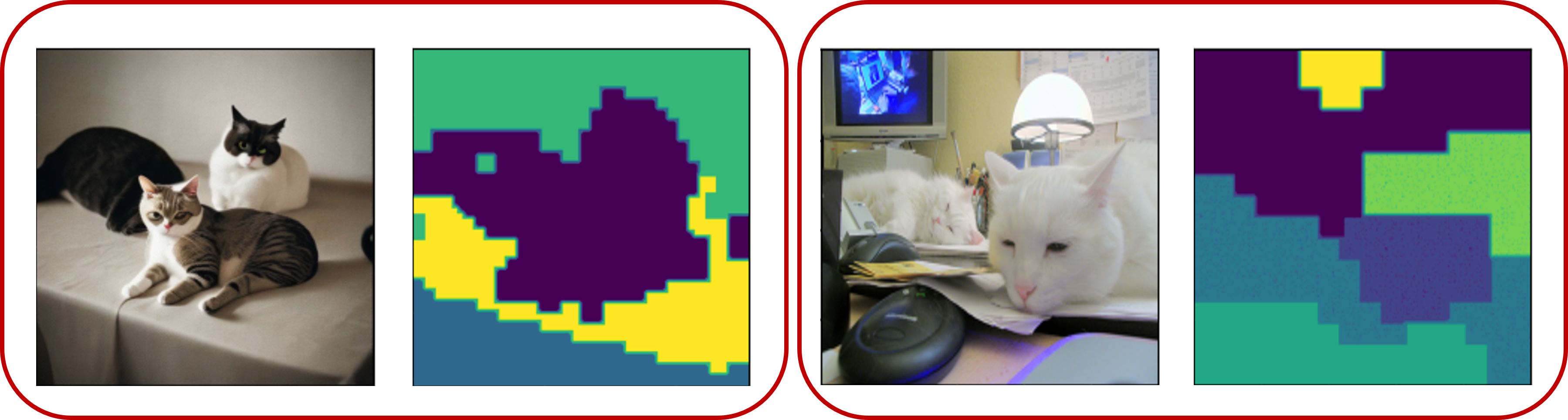}\vspace{-2mm}
  \caption{K-means clustering visualization on UNet features of real image. The image is directly fed into the diffusion model with adding noise once. We use the average feature across the last 6 blocks for visualization.\vspace{-4mm}}
  \label{fig:kmeans_fea}
\end{figure}

%------------------------------------------------------------------------

%-------------------------------------------------------------------------

\section{Utilizing Foundation Model for Downstream Task}
In the first part, we will focus on the typical discriminative foundation model for downstream task. In the second part, we will exploit some current methods utilizing Stable Diffusion for downstream discriminative task. 

\subsection{Visual-Language Model}
Large vision-language model, such as CLIP~\cite{radford2021learning} and ALIGN~\cite{jia2021scaling}, are trained with image-language pairs via contrastive learning, due to its strong zero-shot image recognition performance, there is a new research line dubbed as open-vocabulary objection/segmentation, aiming to introduce the open category recognition ability into the objection or segmentation tasks. Early works on open-vocabulary segmentation such as LSeg~\cite{baranchuk2021label} directly transform the vision-language model classification model to segmentation pipeline. More specifically LSeg directly predicts the category of the pixel embedding with the text embedding, without introducing any extra mask generator module. MaskCLIP~\cite{zhou2022extract} first shows that the value (\textit{V}) embedding output by the CLIP visual part could be used as mask proposal for segmentation, together with the text embedding as the classifier weight the CLIP pipeline could directly output segmentation mask, then it further introduces Mask2Former~\cite{cheng2022masked} to improve the results, which is trained in a self training manner with the predicted segmentation masks. The recent works~\cite{qin2023freeseg,liang2023open} follow the similar pipeline, which typically has two parts: the first part is transformer based mask proposal network and the second part is the CLIP which is to provide open-vocabulary prediction. There are also a few methods elegantly unifying these two parts, for example, ZegCLIP~\cite{zhou2023zegclip} and SAN~\cite{xu2023side} directly adopt CLIP as the main backbone (feature extractor part) and add a lightweight mask generator which takes input feature from CLIP. Since the pipeline with Mask2Former usually takes longer training time, the methods including MaskCLIP (the one without extra mask generator) have fewer parameters and also could achieve better performance, which could be a baseline for future research.

\subsection{Text-to-image Diffusion Model}

Diffusion models are another research hotspot in recent years. The most successful application is text-to-image generation, by fine-tuning or directly utilizing the pretrain diffusion model, where Stable Diffusion~\cite{rombach2022high} is one of the most popular deployed diffusion model. Since the text-to-image generation model\footnote{If not specified, the (text-to-image) diffusion model refers to Stable Diffusion in the subsequent sections.} such as Stable Diffusion is trained with large amount of image-text pairs just like CLIP, a natural question is that whether those cross-modality generative models could be applied to discriminative task? As some pioneer works~\cite{hertz2022prompt} show that the features inside the diffusion model already have rich semantic and localization information, the pretrained diffusion model has potential to be extended to other discriminative tasks.

There are already a few works trying to utilize the text-to-image diffusion model for downstream tasks. Some methods~\cite{clark2023text,li2023your} transform the text-to-image diffusion model to a zero-shot classification model which is competitive to CLIP, by obtaining the posterior classification scores based on the predicted noise during the denoising process. And other methods like OIDSE~\cite{xu2023open} and VPN~\cite{zhao2023unleashing} utilize the UNet features in the diffusion model for downstream tasks such as segmentation and depth estimation. In the following texts we focus on the segmentation task.

In ODISE and VPN, the diffusion model is only to provide features, which will be the input to the subsequent mask generator network such as Mask2Former~\cite{cheng2022masked} or LAVT~\cite{yang2022lavt}, both methods only adopt one time step for the diffusion model, and VPN does not add the noise to the latent vector while ODISE does. In ODISE, an extra learnable module called implicit captioner is proposed to provide the textual embedding to UNet. VPN also utilizes a similar module denoted as text adapter, as well as cross attention maps to be combined with multi-level UNet features. Although these methods achieve good performance in the downstream tasks, we question the efficiency of this naive way of directly using UNet features with one time step. And actually, the ablation study in VPN already shows that there is limited improvement by using the extra text adapter and cross attention features, which indicates that this naive way is not totally efficient to fully exploit the diffusion model for segmentation. This conclusion also holds for the implicit captioner in ODISE.

\begin{table*}[tbp]
  \centering
  \scalebox{0.95}{
  \begin{tabular}{cccccccc}
    \toprule
    \textbf{Method}& \textbf{Architecture} & \textbf{PAS-21} & \textbf{A-847} & \textbf{PC-459} & \textbf{A-150} & \textbf{PC-59} & \textbf{COCO} \\
    \midrule
    MaskCLIP~\cite{ding2022open}&CLIP + Mask2Former & - & 8.2&10.0&23.7&45.9&-\\
    ODISE~\cite{xu2023open} &CLIP + SD + Mask2Former &  84.4& 11.1& 14.5& 29.9& 57.3& 65.2\\
    \hline
   ODISE w/o \textit{IC}&CLIP + SD + Mask2Former &  82.4&10.3&12.4&28.3&54.3&61.6 \\
  ODISE  w/o (\textit{IC} \& \textit{UNet)}&CLIP + SD + Mask2Former& 76.6&10.1&12.4&27.2&51.7&56.3\\
   ODISE * w/o (\textit{IC} \& \textit{VQGAN}) &CLIP + SD + Mask2Former&  80.1&10.2&13.2&28.6&52.5&61.2\\
   \hline
   SAN~\cite{xu2023side} &CLIP + light decoder& 94.6& 12.4&15.7&32.1&57.7&-\\
    \bottomrule
  \end{tabular}}\vspace{-2mm}
  \caption{Detailed ablation study on ODISE, where the model is trained with train set from COCO and the evaluated on other datasets. 'w/o \textit{IC}' denotes not using implicit captioner and instead only adopting null text embedding, \textit{UNet} denotes using features from UNet, and * denotes without adding noise to the output of encoder. We directly use the official ODISE code, due to the limited computation resource, for the three ablation studies we only train 66999, 81999 and 48999 iterations out of the whole 92188 iterations, with 16 A100 GPUs. \vspace{-2mm}}
  \label{tab:odise}
\end{table*}

\section{Experimental Analysis}

In this section, we will first show that the pretrained visual-language model, more specifically the CLIP, has the potential to be directly extended to other downstream tasks. Then, we will show the current methods using text-to-image diffusion model are not efficient with the naive way of deploying pretrained diffusion model as the feature extractor.

\subsection{Visual-Language Model}
We choose the widely used visual-language model CLIP for analysis. We visualize the CLIP visual features under the weakly supervised segmentation task~\cite{lin2023clip}, where every image is provided with its ground-truth class labels. We adopt the Grad-CAM~\cite{selvaraju2017grad} for visualization\footnote{We follow CLIP-ES~\cite{lin2023clip} to use the features before the last attention layer to compute CAM.}. For text prompt input to the CLIP language part, we only use 4 classes here: trees/palms, car, building and windows. The format of text prompt is "a photo of {classname}". The visualization is shown in Fig.~\ref{fig:clip_cam}, it indicates that directly using CLIP features is enough to achieve good localization or segmentation, and also the prompt engineering, \textit{i.e.}, the choice of text prompt, is also important to achieve better results. In Fig.~\ref{fig:clip_cam_threshold}, we further show that simply adopting the binary threshold on the Grad-CAM could lead to refined segmentation. Those findings that CLIP visual features already have localization and semantic information show the potential of the extensibility to other discriminative task. Fully investigating such localization ability of CLIP for segmentation or other tasks is still not widely studied yet in the community.

\subsection{Text-to-image Diffusion Model}

Here we do a detailed analysis for ODISE, which is an open-vocabulary segmentation method based on Stable Diffusion. In ODISE, the image will be fed into the diffusion model with adding noise once, and the features from encoder-decoder in the VQGAN along with the features from the UNet will be used for the subsequent Mask2Former for mask proposal. Unlike the original diffusion model achieving image generation through the denoising of multiple time steps, the UNet feature (with one time step) from the ODISE may have poor quality regarding the semantic and localization information, as a recent method~\cite{patashnik2023localizing} hypothesizes that the denoising process is a coarse-to-fine synthesis with multiple time steps. To verify it, we visualize the cross attention in different scenarios as shown in Fig.~\ref{fig:cross_atten}. In Fig.~\ref{fig:cross_atten}, we first deploy the Stable Diffusion for the normal text-to-image generation with the text prompt 'a horse on the grass', we visualize the cross attention corresponding to the token 'horse' in the last time step, we find these attentions are basically accurate localizing the object. Then we send the generated image back to the diffusion model with adding noise once just like ODISE does, and we also visualize the cross attention of token 'horse'. It turns out that the resulting attention maps become blurry and less accurate for localization, compared to the ones during the generation process. This attention degradation phenomenon may be even more severe if using real image as in the ODISE. Since the UNet features used by ODISE, which will be used by Mask2Former for mask proposal, are directly related to cross attention, the attention degradation may deteriorate the segmentation performance. We also directly visualize the UNet features by k-means clustering in Fig.~\ref{fig:kmeans_fea}, it shows in some case the UNet feature indeed has poor semantic and localization information, as shown in Fig.~\ref{fig:kmeans_fea} (\textit{right}). The finding indicates the necessity of denoising process to get high quality features containing better semantic and localization information.

We also conduct ablation study of ODISE. In ODISE, there are a diffusion model (with adding noise once) and an implicit caption module, the output of which will be utilized as textual/conditional embedding and will be combined with null text embedding via summation. The features from encoder-decoder in VQGAN inside the diffusion model, as well as the features from UNet will be sent to Mask2Former for mask proposal. In Tab.~\ref{tab:odise}, we ablate several modules in ODISE, and it turns out that directly using UNet features with null text embedding (\textit{ODISE w/o IC}) already achieves decent performance, and the performance gain from implicit captioner is relatively limited. Note that not using implicit captioner means only having null text embedding in UNet (unconditional embedding), which is not the right usage way of text-to-image diffusion with conditional and unconditional embeddings, it has not explored the language related information in Stable Diffusion. And in Tab.~\ref{tab:odise}, ODISE, which utilizes CLIP, diffusion model and Mask2Former, is still inferior to SAN, which only uses CLIP and a lightweight decoder network. This indicates the current way using diffusion model in ODISE is relatively naive, and has further space to be improved.

\section{Conclusion}
In this paper, we investigate some recent works on using foundation models for downstream tasks. Features from both discriminative model CLIP and generative model Stable Diffusion, which are trained with large amount of cross-modality data pairs, already contain semantic and localization information, and could be deployed for other discriminative tasks. Although achieving  great performance, the current way using diffusion model for downstream tasks is not efficient. We hope this report could provide some insights for the future research.

\section{Acknowledgement}
We thank the GPUs support from Stability AI.

%%%%%%%%% REFERENCES
{\small
\bibliographystyle{ieee_fullname}
\bibliography{egbib}
}

\end{document}